# Image reconstruction from limited range projections using orthogonal moments

H.Z. Shu[a], J. Zhou[a], G.N. Han[b], L.M. Luo[a], J.L. Coatrieux[c]

[a]*Laboratory of Image Science and Technology, Department of Computer Science and Engineering, Southeast University, 210096, Nanjing, China*

[b]*IRMA, Université Louis Pasteur et C.N.R.S., 7, rue René-Descartes F, 67084 Strasbourg, France*

[c]*Laboratoire Traitement du Signal et de l'Image, Université de Rennes I – INSERM U642, 35042 Rennes, France*

Information about the corresponding author:

Huazhong Shu, Ph.D

Laboratory of Image Science and Technology

Department of Computer Science and Engineering

Southeast University, 210096, Nanjing, China

Tel: 00-86-25-83 79 42 49

Fax: 00-86-25-83 79 26 98

Email: shu.list@seu.edu.cn

**Abstract**: A set of orthonormal polynomials is proposed for image reconstruction from projection data. The relationship between the projection moments and image moments is discussed in detail, and some interesting properties are demonstrated. Simulation results are provided to validate the method and to compare its performance with previous works.

1. **Introduction**

In its classical formulation, computerized tomography (CT) deals with the reconstruction of an object from measurements which are line integrals of that object at some known orientations. This formulation has found many applications in the fields of medical imaging, synthetic aperture radar, electron-microscopy based tomography, etc [1]. The Radon transformation, due to its explicit geometric meaning, has played an important role.

In the past decades, a number of studies have been carried out on this subject. Lewitt has summarized a series of projection theorems [2], a detailed description about the properties of Radon transform and its relationship to other transforms have been given by Deans [1]. Major contributions have been reported in the biomedical engineering literature, one of the most active fields for the reconstruction problem, in order to study new X-ray source-detector trajectories (typically a spiral for CT) and to deal with truncated [3] and cone-beam projections [4].

Recently, the moment-based approaches to tomographic reconstruction have attracted considerable attention of several research groups. Salzman [5] and Goncharev [6] respectively proposed the methods based on the moments to find the view angle from the projection data. Milanfar et al. [7] described a variational framework for the tomographic reconstruction of an image from the maximum likelihood estimates of its orthogonal moments. Basu and Bresler [8, 9] discussed the problem of recovering the view angles using moments of the projections. Wang and Sze [10] proposed an approach based on the relationship between the projection moments and the image moments to reconstruct the CT images from limited range projections. In the Wang's algorithm, the geometric moments were used and interesting results have been obtained. However, the use of the geometric moments has the following disadvantages: (1) the geometric moments of an image are integrals of the field shape over space, and the image can be uniquely determined by the geometric moments of all orders. They are sensitive to digitization error and minor shape deformations [11]; (2) the geometric moments are basically projections of the image function onto the monomials $x^n y^m$. Unfortunately, the basis set $\{x^n y^m\}$ is not orthogonal. These moments are therefore not optimal with regard to the information redundancy and other useful properties that may result from using orthogonal basis functions.

To overcome these inconvenients, we propose in this paper, by extending Wang's algorithm, a new moment-based approach using the orthogonal basis set to reconstruct the image from limited range projection.

The paper is organized as follows. A brief review of Radon transform and the definition of projection moments and image moments are given in Section 2. In Section 2, we also establish the relationship between projection moments and image moments and discuss how to estimate the projection moments at any specific view from image moments. Simulation results are provided in Section 3. Section 4 concludes the paper and provides some additional perspectives.

## 2. Method

We first sketch the basics of Radon transform. The orthogonal projection moments, defined in terms of normalized polynomials, are then introduced in subsection 2.2. Some theorems relating projection and image moments are reported and demonstrated in subsection 2.3. In the last subsection, we show how to estimate the unknown projections from the calculated image moments.

*2.1 Radon transform*

Let $f(x, y) \in L^2(D)$ be a square-integrable function with support inside the unit circle $D$ in the plane and $g(s, \theta)$ be the Radon transform of $f(x, y)$ defined as follows

$$g(s,\theta) = \iint_D f(x,y)\delta(x\cos\theta + y\sin\theta - s)dxdy \tag{1}$$

where $\delta(\cdot)$ denotes the Dirac delta function, $s$ is the distance from the origin to the ray, and $\theta$ is the angle between the x-axis and the ray.

For a given view $\theta$, the two-dimensional (2D) function $g(s, \theta)$ becomes a one-variable of $s$, denoted by $g_\theta(s)$. Since $g_\theta(s)$ represents a collection of integrals along a set of parallel rays, it is also called parallel projection of $g(s, \theta)$ at view $\theta$ [10]. The Radon transform given by Eq. (1) can be rewritten as

$$g_\theta(s) = \iint_D f(x,y)\delta(x\cos\theta + y\sin\theta - s)dxdy \tag{2}$$

*2.2 Orthogonal projection moments and image moments*

The moments of $g_\theta(s)$ are called projection moments in the Radon domain [2]. In this paper, we use a set of orthonormal polynomials instead of the set of monomials $\{s^p\}$ to define the projection moments. Let $\{P_p(s)\}$, $p = 0, 1, 2, \ldots \infty$, be a set of orthonormal polynomials defined on the interval $[-1, 1]$, the $p$th order orthonormal projection moment of $g_\theta(s)$ is defined as

$$L_p(\theta) = \int_{-1}^{1} P_p(s)g_\theta(s)ds \tag{3}$$

Let $\lambda_{nm}$ be the $(n + m)$th order orthogonal moment of $f(x, y)$ defined as [12]

$$\lambda_{nm} = \iint_D P_n(x)P_m(y)f(x,y)dxdy \tag{4}$$

Substitution of Eq. (2) into (3) yields

$$L_p(\theta) = \int_{-1}^{1}\iint_D f(x,y)\delta(x\cos\theta + y\sin\theta - s)P_p(s)dxdyds \quad (5)$$
$$= \iint_D \int_{-1}^{1} f(x,y)\delta(x\cos\theta + y\sin\theta - s)P_p(s)dsdxdy$$

Using the property of delta function, we have

$$L_p(\theta) = \iint_D P_p(x\cos\theta + y\sin\theta)f(x,y)dxdy \quad (6)$$

This last equation will allow us to establish a relationship between the orthogonal projection moments defined by Eq. (3) and the orthogonal moments of $f(x, y)$ defined by Eq. (4). This is the objective of the next subsection.

*2.3 Relationship between projection moments and image moments*

Let us first introduce some basic definitions. Let the $p$th order normalized polynomials $P_p(t)$ be

$$P_p(t) = \sum_{r=0}^{p} c_{pr} t^r \quad (7)$$

and let $V_p(t) = (P_0(t), P_1(t), P_2(t), ..., P_p(t))^T$ and $M_p(t) = (1, t, t^2, ..., t^p)^T$ where the subscript $T$ indicates the transposition, then we have

$$V_p(t) = C_p M_p(t) \quad (8)$$

where $C_p = (c_{kr})$, with $0 \le r \le k \le p$, is a $(p+1) \times (p+1)$ lower triangular matrix.

Since all the diagonal elements of $C_p$, $c_{kk}$, are not zero, the matrix $C_p$ is non-singular, thus

$$M_p(t) = C_p^{-1} V_p(t) = D_p V_p(t) \quad (9)$$

where $D_p = (d_{kr})$, with $0 \leq r \leq k \leq p$, is the inverse matrix of $C_p$.

Eq. (9) can be rewritten as

$$t^k = \sum_{r=0}^{k} d_{kr} P_r(t), \quad \text{for } 0 \leq k \leq p \tag{10}$$

We then have the following theorem about the projection moments and image moments:

**Theorem 1**. The orthogonal projection moment of order $p$ at given view $\theta$, $L_p(\theta)$, can be expressed as a linear combination of image moments of same order and lower, i.e.,

$$L_p(\theta) = \sum_{n=0}^{p} \sum_{m=0}^{p-n} \mu_{nm}(p,\theta) \lambda_{nm} \tag{11}$$

where

$$\mu_{nm}(p,\theta) = \sum_{q=0}^{p-(n+m)} \sum_{r=0}^{q} c_{p,q+n+m} d_{n+r,n} d_{m+q-r,m} \binom{n+m+q}{n+r} (\cos\theta)^{n+r} (\sin\theta)^{m+q-r} \tag{12}$$

The proof of Theorem 1 is deferred to Appendix A.

Theorem 1 can also be expressed in matrix form. To do this, let us introduce the notations: $\Phi_M(\theta) = [L_0(\theta), L_1(\theta), ..., L_M(\theta)]^T$, $\lambda^{(k)} = [\lambda_{k0}, \lambda_{k-1,1}, ..., \lambda_{1,k-1}, \lambda_{0k}]^T$, and $\Psi_M = [\lambda^{(0)T}, \lambda^{(1)T}, ..., \lambda^{(M)T}]^T$ where $M$ denotes the maximum order of moments we want to utilize. Then, we have

$$\Phi_M(\theta) = T_M(\theta) \Psi_M \tag{13}$$

where $T_M(\theta)$ is a matrix of size $(M+1) \times (M+1)(M+2)/2$ whose definition is given by

$$T_M(\theta) = \begin{pmatrix} \mu_{00}(0,\theta) \\ \mu_{00}(1,\theta) & \mu_{10}(1,\theta) & \mu_{01}(1,\theta) \\ \mu_{00}(2,\theta) & \mu_{10}(2,\theta) & \mu_{01}(2,\theta) & \mu_{20}(2,\theta) & \mu_{11}(2,\theta) & \mu_{02}(2,\theta) \\ \ldots\ldots \\ \mu_{00}(M,\theta) & \mu_{10}(M,\theta) & \mu_{01}(M,\theta) & \mu_{20}(M,\theta) & \mu_{11}(M,\theta) & \mu_{02}(M,\theta) & \cdots \end{pmatrix} \quad (14)$$

Here $\mu_{nm}(p, \theta)$, with $0 \leq n, m, p \leq M$, are defined by Eq. (12).

Theorem 1 not only provides the solution for finding the projection moments from image moments, but also the solution for finding image moments from projection moments. Concerning this latter point, we have the following proposition:

**Proposition 1**. Given line integral projections of $f(x, y)$ at $m$ different angles $\theta_i$ in $[0, \pi)$, $1 \leq i \leq m$, one can uniquely determine the first $m$ moment vectors $\lambda^{(k)}$ of $f(x, y)$, $0 \leq k < m$. This can be done using only the first $m$ orthogonal moments $L_p(\theta_i)$, $0 \leq p < m$ of the projections.

For the proof of Proposition 1, we refer to Milanfar et al. [7].

Note that Theorem 1 is valid for any type of polynomials and therefore the reconstruction based on geometric moments is a particular case of the problem considered here. In fact, for the geometric moments, the coefficients $c_{kr}$ and $d_{kr}$ in Eqs. (7) and (10) are given by

$$c_{kr} = d_{kr} = \delta_{kr} \quad (15)$$

where $\delta_{kr}$ is the Kronecker symbol.

In this case, Eq. (12) becomes

$$\mu_{nm}(p,\theta) = \binom{n+m}{n} c_{p,n+m} (\cos\theta)^n (\sin\theta)^m = \binom{p}{n} (\cos\theta)^n (\sin\theta)^{p-n} \quad (16)$$

In the remaining part of the paper, we focus on the use of orthogonal polynomials. In particular, we are interested in the normalized Legendre polynomials. The $p$th order normalized Legendre polynomial $P_p(x)$ over $[-1, 1]$ is defined by

$$P_p(x) = \sqrt{\frac{2p+1}{2}} \frac{1}{2^p p!} \frac{d^p}{dx^p}(x^2-1)^p = \sum_{r=0}^{p} c_{pr} x^r \qquad (17)$$

with

$$c_{pr} = \begin{cases} (-1)^{(p-r)/2} \sqrt{\frac{2p+1}{2}} \frac{1}{2^p} \frac{(p+r)!}{[(p-r)/2]![(p+r)/2]! r!}, & \text{for } p-r \text{ even} \\ 0, & \text{for } p-r \text{ odd} \end{cases} \qquad (18)$$

An essential step when applying a special polynomial to the reconstruction problem is to find the inverse matrix $D_p$ in Eq. (9). For the normalized Legendre polynomials, we have

**Proposition 2**. For the lower triangular matrix $C_p$ whose elements $c_{kr}$ are defined by Eq. (18), the elements of the inverse matrix $D_p$ are given as follows

$$d_{kr} = \begin{cases} \sqrt{\frac{2}{2r+1}} \frac{2^{\frac{3}{2}r - \frac{1}{2}k}}{[(k-r)/2]! \prod_{j=1}^{(k-r)/2}(2r+2j+1)} \frac{k! r!}{(2r)!}, & \text{for } k-r \text{ even} \\ 0, & \text{for } k-r \text{ odd} \end{cases}, \quad 0 \le r \le k \le p \quad (19)$$

The proof of Proposition 2 is deferred to Appendix A.

Based on the above proposition, we are now ready to give an explicit expression of $\mu_{nm}(p, \theta)$ in Eq. (12) for orthonormal Legendre polynomials. Let

$$v_{qr}(p,n,m) = c_{p,q+n+m} d_{n+r,n} d_{m+q-r,m} \binom{n+m+q}{n+r} \qquad (20)$$

from Eq. (19), it is obvious that

$$v_{qr}(p,n,m) = 0 \text{ for } q \text{ or } r \text{ being odd number} \tag{21}$$

Eq. (12) can thus be rewritten as

$$\mu_{nm}(p,\theta) = \sum_{q=0}^{p-(n+m)} \sum_{r=0}^{q} v_{qr}(p,n,m)(\cos\theta)^{n+r}(\sin\theta)^{m+q-r} \tag{22}$$

where the notation $\sum_{q=0}^{p-(n+m)}$ stands for the summation with respect to the even value of $q$ varying from 0 to $p - (n+m)$.

Substitution of Eqs. (18) and (19) into (20) yields

$$v_{qr}(p,n,m) = \sqrt{\frac{2(2p+1)}{(2n+1)(2m+1)}} \frac{(-1)^{[p-(n+m+q)]/2}}{2^{p-n-m+q/2}} \frac{1}{\prod_{i=1}^{r/2}(2n+2i+1)\prod_{j=1}^{(q-r)}(2m+2j+1)}$$
$$\times \frac{n!m!}{(2n)!(2m)!} \frac{(p+n+m+q)!}{(r/2)![(q-r)/2][(p-n-m-q)/2][(p+n+m+q)/2]!} \tag{23}$$

The apparent complexity of Eq. (23) can be omitted by the following recurrence relations

$$v_{q+2,r}(p,n,m) = -\frac{(p+n+m+q+1)(p-n-m-q)}{(q-r+2)(2m+q-r+3)} v_{qr}(p,n,m) \tag{24}$$

$$v_{q,r+2}(p,n,m) = -\frac{(q-r)(2m+q-r+1)}{(r+2)(2n+r+3)} v_{qr}(p,n,m) \tag{25}$$

$$v_{00}(p,n+2,m) = -\sqrt{\frac{1}{(2n+1)(2n+5)}} \frac{(p-n-m)(p+n+m+1)}{2n+3} v_{00}(p,n,m) \tag{26}$$

$$v_{00}(p,n+1,m-1) = \sqrt{\frac{(2m-1)(2m+1)}{(2n+1)(2n+3)}} v_{00}(p,n,m) \tag{27}$$

$$v_{00}(p,n,m) = \sqrt{\frac{2(2p+1)}{(2n+1)(2m+1)}} \frac{(-1)^{(p-n-m)/2}}{2^{p-n-m}} \frac{n!m!}{(2n)!(2m)!} \frac{(p+n+m)!}{[(p-n-m)/2][(p+n+m)/2]!}$$
$$\tag{28}$$

*2.4 Estimation of the projection moments at any view*

In real situations, projections from certain views are known while projections from other views are unknown. To distinguish them, we adopt here the same terminology used in [10]. The projections are called given projections if the views are given views; the unknown projections correspond to the projections whose views are unknown. In the previous subsection, we have discussed the relationship between the given projection moments and the image moments. The following shows how to compute the unknown projections from the image moments.

When the orthonormal polynomials $P_p(s)$ are used in Eq. (3), the orthogonality property leads to the following inverse transform

$$g_\theta(s) = \sum_{p=0}^{\infty} L_p(\theta) P_p(s) \tag{29}$$

Eq. (29) calculates $g_\theta(s)$, or $g(s, \theta)$, from its projection moments. If only the projection moments of order up to $M$ are used, Eq. (29) is approximated by

$$\tilde{g}_\theta(s) = \sum_{p=0}^{M} L_p(\theta) P_p(s) \tag{30}$$

Substituting Eq. (11) into (30), we have

$$\tilde{g}_\theta(s) = \sum_{p=0}^{M} P_p(s) \sum_{n=0}^{p} \sum_{m=0}^{p-n} \mu_{nm}(p,\theta) \lambda_{nm} \tag{31}$$

where $\mu_{nm}(p, \theta)$ are defined by Eq. (12). Note that when the orthonormal Legendre polynomials are used, Eqs. (22) and (23) allow to compute $\mu_{nm}(p, \theta)$.

The above formula establishes the connection between projections from any specific view and image moments. To summarize, the projections of unknown views can be estimated from given projections as follows

1) Compute the image moments of order up to $M$ from given $M+1$ views;

2) Calculate the unknown projections from the image moments of order up to $M$ by using Eq. (31).

## 3. Results

*3.1 Simulations*

We first apply the proposed solution to a simulated phantom $f(x, y)$, shown in Fig. 1, in order to compare the two methods for computing the image moments. The first method is directly based on Eq. (4) since the image is known, the orthonormal Legendre moments, $\lambda_{nm}$, of order up to 15 are calculated using Eq. (4). The other one is achieved through the computation of projection moments. A projection simulation program is used to generate all the projections from 15 given views and the image moments can then be computed using Proposition 1. The moment values using these two methods are shown in Table 1, where the values without parentheses are computed from the image itself, and the values with parentheses are obtained from projection moments. It can be seen from this table that the errors between the real image moments and calculated image moments are small.

When the image moments of order up to $M$ are calculated, Eq. (31) is used to estimate the projection moments, $\widetilde{g}_{\theta}(s)$, at any specified view $\theta$. In order to test the robustness of the proposed method, we first consider the case where all the projections are known in the interval [0, $\pi$). The estimation of $\tilde{g}_\theta(s)$ from the image moments with maximum order $M = 15$ at views of $0^0$, $15^0$, $60^0$, $90^0$, $120^0$ and $165^0$ is shown in Fig. 2. For comparison, the real projections (computed by a projection simulation program) and the estimation results from these views using the geometric moment-based method proposed by Wang and Sze [10] are also shown in this figure. Fig. 2 shows that the results obtained by the proposed method and Wang's algorithm are almost the same. We then assume that projections are available in the range of $20^0 \leq \theta \leq 160^0$, but unknown in the ranges $0^0 \leq \theta < 20^0$ and $160^0 < \theta \leq 179^0$. The results using both the proposed method and Wang's method are illustrated in Fig. 3. It can be observed that our method performs better than Wang's method, especially at views $0^0$, $15^0$ and $165^0$.

In the second example, we consider the image reconstruction problem from limited range projections. The filtered backprojection (FBP) algorithm is used in the reconstruction process. Since this algorithm requires the projection, $g_\theta(s)$ or $g(s, \theta)$, for all $s$ and all $\theta$ in the whole interval [0, $\pi$), we use the image moment method to estimate the unknown projections. The original grey level image of size $128 \times 128$ is depicted in Fig. 4(a) (it is the same as that used in the previous example), the projections being assumed known from $25^0$ to $155^0$. A projection simulation program is applied to compute the given projection in which the following configuration is used: (a) the total view varies from $0^0$ to $179^0$; (b) the angular sampling rate is $1^0$ per view; (c) the ray sampling rate is 128 rays per view. Fig. 4(b) shows the reconstructed image from incomplete projections ($25^0$-$155^0$) using the FBP method.

We then use the method proposed in the previous section as well as the Wang's method for calculating the image moments with maximum order $M$ equaling to 5, 10, 15, and 20, respectively. These moments are then used to estimate the unknown projections. Based on the given projections and the estimated projections ($0^0$-$25^0$ and $155^0$-$179^0$ for this example), the FBP method is applied to reconstruct the image. The reconstructed images based on both the proposed method and Wang's method are illustrated in Fig. 4(c)-(i). It can be visually seen that our method leads to better results for the same value of $M$. Note that the quality of the reconstructed image for Wang's method with $M = 20$ is poor due to the numerical instability, so that we have not shown this result in Fig. 4. We also use the mean square error (MSE) to qualitatively measure the difference between the original image and the reconstructed image. The MSE is defined as follows

$$MSE(\%) = \frac{\left\|f^* - f\right\|^2}{\left\|f\right\|^2} \times 100\% \tag{32}$$

where $f$ is the original image, and $f^*$ is the reconstructed image.

Table 2 shows the MSEs for the proposed method and Wang's method with different values of $M$. The values of MSE again demonstrate the better performance of the proposed method compared to Wang's method.

*3.2 Evaluation on "pseudo-real" data*

The projection data are not made available by the imaging manufacturers, thus the experimentation has been conducted using a reconstructed Computed Tomography image depicted Fig 5(a). The image intensity values (e.g. density levels) have been inverted in order to enhance the internal details of the brain slice and the tumor features. We have applied a two steps procedure: (i) the projection set has been computed by using a parallel geometry (thus the so-called "pseudo-real" data) over the limited angular range; (ii) the above reconstruction algorithms have then been used.

In this example, the projections are assumed to be known on the interval $[\alpha, 180^0 - \alpha]$ where $\alpha$ is an adjustable parameter. We apply the proposed method to compute the image moments of order up to $M = 25$ for $\alpha$ equaling to $5^0$, $10^0$, $15^0$, $20^0$, $25^0$ and $30^0$, respectively. These moments are then used to estimate the unknown projections according to (31). The reconstructed images based on both the proposed method and FBP method are depicted in Fig. 5(b)-(m). Table 3 shows the corresponding MSEs for both methods. Note that in the images reconstructed using FBP method (Fig. 5(b)-(g)), the image intensity value of the unknown projections is assigned to 0. It can be observed from Fig. 5(h)-(m) that the proposed method leads to good results even for large $\alpha$.

In order to compare the performance of the proposed method with Wang's method, we calculate the image moments of order up to $M = 15$ for different values of $\alpha$ ($\alpha = 5^0$, $10^0$, $15^0$, $20^0$, $25^0$ and $30^0$); these moments are then used to estimate the unknown projections. The MSEs of the reconstructed imaged using FBP method for both methods are illustrated in Fig. 6. Fig. 6 points out the better behavior of our method over the geometric moment-based method presented in [10] in terms of the reconstruction capability from limited range projections.

## 4. Discussion and Conclusion

In this paper, by extending Wang's method, we have proposed a new approach based on the orthogonal moments to solve the tomographic reconstruction problem from limited range projections. By demonstrating some properties of Legendre polynomials, we have established the relationship between the orthogonal projection moments and orthogonal image moments. This relationship permits us to estimate the unknown projections from the computed image moments so that the reconstruction problem from limited range projections can be well solved.

It has been shown that orthogonal moments have some merits in comparison with the geometric moments. First, the geometric moments, especially at high order, are sensitive to noise and digitization error. Second, the orthogonal moments have simple inverse transform, thus, the image can be more easily reconstructed from the orthogonal moments. This new approach was evaluated through a fully known phantom and a sample of "pseudo-real" data set. Experimental results have shown that our method is efficient, and provides better reconstruction results compared to Wang's method.

Reconstruction from limited range projections is of relevance for a number of medical imaging modalities, in particular for X-ray devices. Systems such as Electron Beam Computed Tomography (EBCT) and Rotational-X (Rot-X) make use of a reduced set of projections and allow decreasing the radiation dose for the patient, an important health care constraint. However, the use of sparse data leads to more visible reconstruction artifacts. A compromise is then required in order to preserve the relevant information. We have not addressed in this paper the issue regarding the minimal number of range projections from which a reasonable reconstruction could be obtained using the above methods. This is a difficult problem because it can not be addressed by MSE computation alone; subjective analysis by physicians in order to evaluate the medical consequence of a loss of informational contents is also of major importance. It also depends on the targeted use: when diagnosis requires high reconstruction performance, interventions may accept lower precision in reconstruction as far as faster algorithms are available. The works in progress are therefore devoted to such issues and also the comparison with other reconstruction algorithms (in complexity for instance) and the performance that can be expected for vascular reconstruction when dealing with a modality like Rot X. They also focus on the extension of the method to other projection modes such as cone-beam projections.

*Acknowledgements:* This work was supported by National Basic Research Program of China under grant, No. 2003CB716102, the National Natural Science Foundation of China under grant No. 60272045 and Program for New Century Excellent Talents in University under grant No. NCET-04-0477.

**Appendix A**

**Proof of Theorem 1**. Using Eq. (7), Eq. (6) can be rewritten as

$$L_p(\theta) = \iint_D f(x,y) \sum_{q=0}^{p} c_{pq} (x\cos\theta + y\sin\theta)^q \, dxdy$$

$$= \iint_D f(x,y) \sum_{q=0}^{p} \sum_{r=0}^{q} c_{pq} \binom{q}{r} (\cos\theta)^r (\sin\theta)^{q-r} x^r y^{q-r} \, dxdy \quad (A1)$$

where $\binom{q}{r} = \dfrac{q!}{r!(q-r)!}$ is the combination number.

Using Eq. (10), we have

$$L_p(\theta) = \iint_D f(x,y) \sum_{q=0}^{p} \sum_{r=0}^{q} c_{pq} \binom{q}{r} (\cos\theta)^r (\sin\theta)^{q-r} \sum_{n=0}^{r} \sum_{m=0}^{q-r} d_{rn} d_{q-r,m} P_n(x) P_m(y) \, dxdy$$

$$= \sum_{q=0}^{p} \sum_{r=0}^{q} c_{pq} \binom{q}{r} (\cos\theta)^r (\sin\theta)^{q-r} \sum_{n=0}^{r} \sum_{m=0}^{q-r} d_{rn} d_{q-r,m} \iint_D P_n(x) P_m(y) f(x,y) \, dxdy \quad (A2)$$

$$= \sum_{n=0}^{p} \sum_{m=0}^{p-n} \sum_{q=n+m}^{p} \sum_{r=n}^{q-m} c_{pq} \binom{q}{r} (\cos\theta)^r (\sin\theta)^{q-r} d_{rn} d_{q-r,m} \iint_D P_n(x) P_m(y) f(x,y) \, dxdy$$

By using Eq. (4) and by changing the variables $q = q' + n + m$ and $r = r' + n$ in the last equation of the above expression, we obtain

$$L_p(\theta) = \sum_{n=0}^{p} \sum_{m=0}^{p-n} \sum_{q=0}^{p-(n+m)} \sum_{r=0}^{q} c_{p,q+n+m} d_{r+n,n} d_{q-r+m,m} \binom{q+n+m}{r+n} (\cos\theta)^{r+n} (\sin\theta)^{q-r+m} \lambda_{nm} \quad (A3)$$

The proof is now achieved. □

**Proof of Proposition 2**. To prove the proposition, we need to demonstrate the following relation

$$\sum_{r=l}^{k} c_{kr} d_{rl} = \delta_{kl} \qquad \text{for } 0 \leq k, l \leq p \tag{A4}$$

From Eqs. (18) and (19), it is obvious that Eq. (A4) is true when the two integers $k$ and $l$ have different parity. Thus, we need only to demonstrate it for the following two cases: (1) $k$, $r$ and $l$ are all even numbers; (2) $k$, $r$ and $l$ are all odd numbers. We consider the first case. Let $k = 2u$, $r = 2v$ and $l = 2w$, we have

$$c_{kr} = c_{2u,2v} = \sqrt{\frac{4u+1}{2}} \frac{(-1)^{u-v}}{2^{2u}} \frac{(2u+2v)!}{(u-v)!(u+v)!(2v)!} \tag{A5}$$

$$d_{rl} = d_{2v,2w} = \sqrt{\frac{2}{4w+1}} \frac{2^{3v-w}}{(v-w)! \prod_{j=1}^{(v-w)}(4w+2j+1)} \frac{(2v)!(2w)!}{(4w)!}$$

$$= \sqrt{2(4w+1)} \frac{2^{2w+1}(2v)!(v+w+1)!}{(v-w)!(2v+2w+2)!} \tag{A6}$$

thus

$$\sum_{r=l}^{k} c_{kr} d_{rl} = \sum_{2v=2w}^{2u} c_{2u,2v} d_{2v,2w} = (-1)^{u} 2^{2(w-u)+1} \sqrt{(4u+1)(4w+1)} \sum_{v=w}^{u} S(u,w,v) \tag{A7}$$

where

$$S(u,w,v) = \frac{(-1)^{v}(2u+2v)!(v+w+1)!}{(u-v)!(u+v)!(v-w)!(2v+2w+2)!} \tag{A8}$$

Let

$$G(u,w,v) = \frac{(-1)^{v+1}}{2} \frac{(2u+2v)!(v+w)!}{(u+1-v)!(u+v)!(v-w)!(2v+2w)!} \frac{(u+1-v)(v-w)}{(2u+2w+1)(u-w)},$$

$$\text{for } w < u \tag{A9}$$

it can be verified

$$S(u,w,v) = G(u,w,v+1) - G(u,w,v), \text{ for } w \leq v \leq u \text{ and } w < u \tag{A10}$$

thus

$$\sum_{v=w}^{u} S(u,w,v) = \sum_{v=w}^{u} [G(u,w,v+1) - G(u,w,v)] = G(u,w,u+1) - G(u,w,w) \quad (A11)$$

It can be easily deduced from Eqs. (A9) and (A11) that

$$\sum_{v=w}^{u} S(u,w,v) = 0 \quad \text{for } w < u \quad (A12)$$

When $k = l$, i.e., $u = w$, we have

$$c_{2u,2u} d_{2u,2u} = 2(4u+1) \frac{(4u)!(2u+1)!}{(2u)!(4u+2)!} = 1 \quad (A13)$$

The proof of the proposition when $k$, $r$ and $l$ are even numbers is now complete. For $k$, $r$ and $l$ being odd numbers, the proposition can be demonstrated in a similar way and is not given here. □

Note that the proof of Proposition 1 was inspired by a technique proposed by Zeilberger [13].

**About the Author**—HUAZHONG SHU received the B. S. Degree in Applied Mathematics from Wuhan University, China, in 1987, and a Ph. D. degree in Numerical Analysis from the University of Rennes (France) in 1992. He was a postdoctoral fellow with the Department of Biology and Medical Engineering, Southeast University, from 1995 to 1997. He is now with the Department of Computer Science and Engineering of the same university. His recent work concentrates on the treatment planning optimization, medical imaging, and pattern recognition. Dr. SHU is a member of IEEE.

**About the Author**—JIAN ZHOU received the B. S. Degree and M. S. Degree both in Radio Engineering from Southeast University, China, in 2003. He is now a Ph. D student of the Laboratory of Image Science and Technology of Southeast University. His current research is mainly focused on image processing.

**About the Author**—GUONIU HAN received the B. S. Degree in Applied Mathematics from Wuhan University, China, in 1987, and a Ph. D. degree in Mathematics from the University of Strasbourg I (France) in 1992. He works as Research Associate (CR) at French National Center for Scientific Research (CNRS) since 1993. His recent work focuses on the algebraic combinatorics, computer algebra and pattern recognition.

**About the Author**—LIMIN LUO obtained his Ph. D. degree in 1986 from the University of Rennes (France). Now he is a professor of the Department of Computer Science and Engineering, Southeast University, Nanjing, China. He is the author and co-author of more than 80 papers. His current research interests include medical imaging, image analysis, computer-assisted systems for diagnosis and therapy in medicine, and computer vision. Dr LUO is a senior member of the IEEE. He is an associate editor of *IEEE Eng. Med. Biol. Magazine* and *Innovation et Technologie en Biologie et Medecine* (ITBM).

**About the Author**—JEAN-LOUIS COATRIEUX received the PhD and State Doctorate in Sciences in 1973 and 1983, respectively, from the University of Rennes 1, Rennes, France. Since 1986, he has been Director of Research at the National Institute for Health and Medical Research (INSERM), France, and since 1993 has been Professor at the New Jersey Institute of Technology, USA. He is also Professor at Telecom Bretagne, Brest, France. He has been the Head of the Laboratoire Traitement du Signal et de l'Image, INSERM, up to 2003. His experience is related to 3D images, signal processing, pattern recognition, computational modeling and complex systems with applications in integrative biomedicine. He published more than 300 papers in journals and conferences and edited many books in these areas. He has served as the Editor-in-Chief of the IEEE Transactions on Biomedical Engineering (1996-2000) and is in the Boards of several journals. Dr. COATRIEUX is a fellow member of IEEE. He has received several awards from IEEE (among which the EMBS Service Award, 1999 and the Third Millennium Award, 2000) and he is Doctor Honoris Causa from the Southeast University, China.

Table 1: Comparison of image moment values obtained from Eq. (4) (without parentheses) and image moment values estimated from projection moments (with parentheses)

| n \ m | 0 | 2 | 5 | 7 |
|---|---|---|---|---|
| 0 | 0.80200 (0.80200) | -0.59407 (-0.59396) | -0.079439 (-0.079385) | 0.037834 (0.037785) |
| 2 | -0.67940 (-0.67929) | 0.47723 (0.47706) | -0.086657 (-0.086584) | -0.052171 (0.052094) |
| 4 | 0.33773 (0.33757) | -0.19324 (-0.19310) | -0.082248 (-0.082153) | 0.049279 (0.049188) |
| 8 | 0.059616 (0.059581) | -0.02436 (-0.02436) | -0.065381 (-0.065233) | 0.038325 (0.038214) |

Table 2: Comparison of reconstruction MSE (%) using geometric moment method and orthonormal moment method

| Moment order | 5 | 10 | 15 | 20 |
|---|---|---|---|---|
| MSE (Geometric) | 11.7358 | 9.8997 | 27.4587 | -- |
| MSE (Legendre) | 11.7344 | 9.8863 | 6.8392 | 6.5655 |

Table 3: Comparison of reconstruction MSE (%) using FBP method and orthonormal moment method with maximum order $M = 25$

| Value of $\alpha$ | 5 | 10 | 15 | 20 | 25 | 30 |
|---|---|---|---|---|---|---|
| MSE (FBP) | 2.3462 | 7.0099 | 13.6023 | 22.1117 | 32.7264 | 45.7178 |
| MSE (Legendre) | 0.8022 | 1.4430 | 2.1014 | 2.7697 | 3.4405 | 4.0936 |

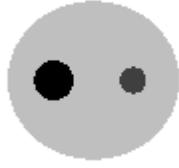

Figure 1: Computer simulated phantom.

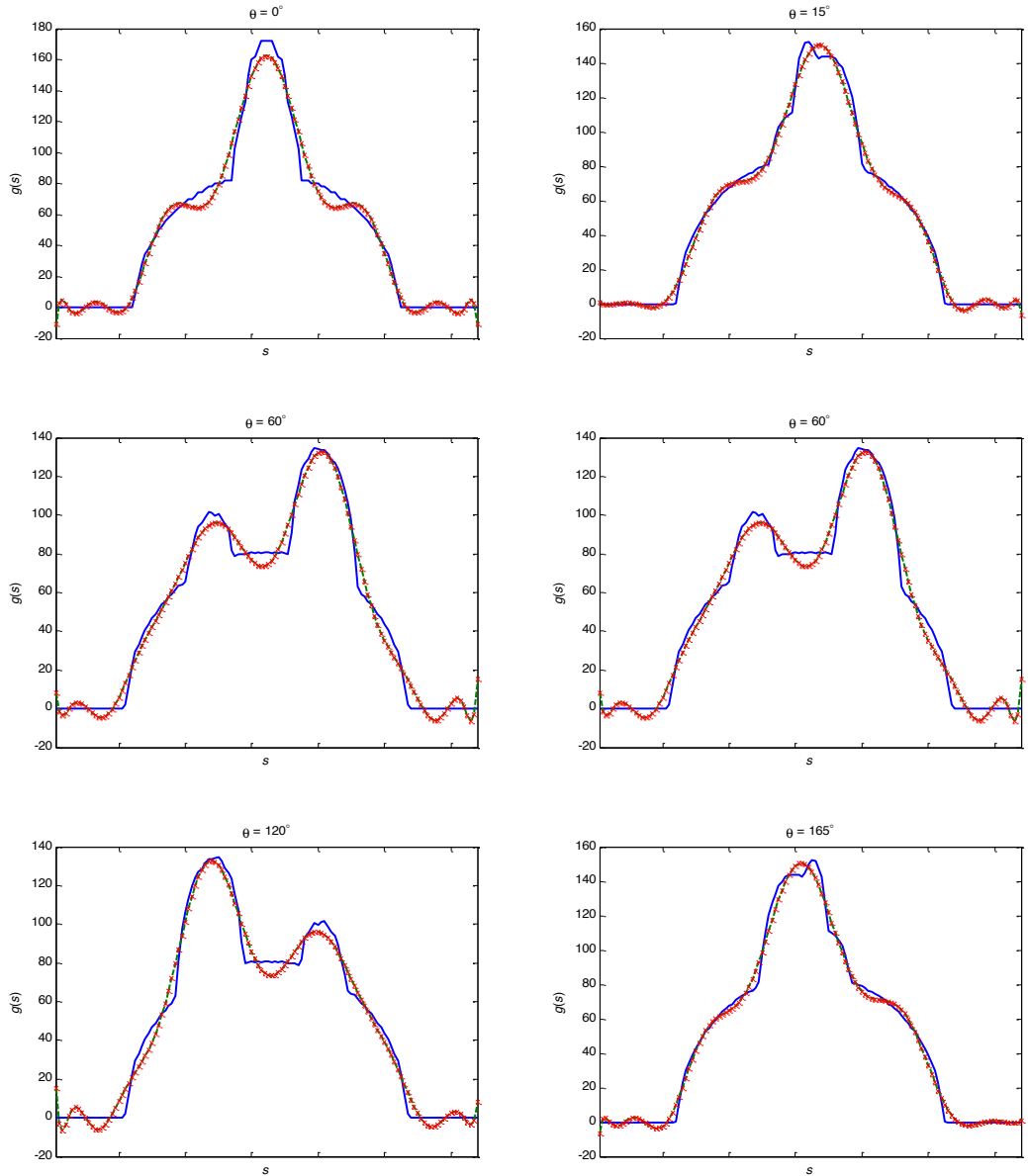

Figure 2: Results when all the projections are known. The projections are compared at different views: $0^0$, $15^0$, $60^0$, $90^0$, $120^0$, $165^0$. Image moments of order up to 15 are used. (solid line: original projections; dash line: projections estimated from Legendre moments; cross: projections estimated from geometric moments)
26

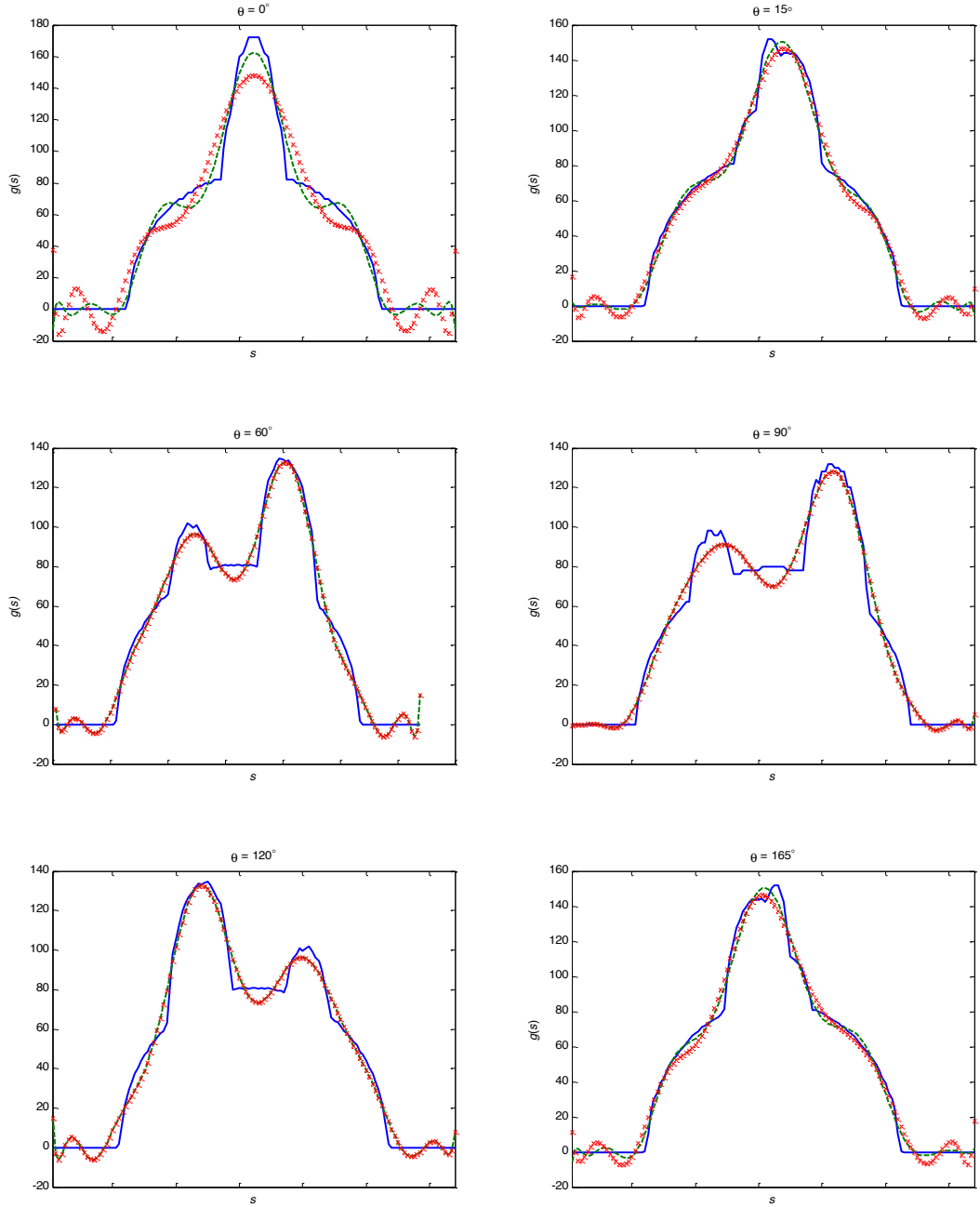

Figure 3: Case where the projections are available at the limited range ($20^0$-$160^0$). Projected data are compared at different views: $0^0$, $15^0$, $60^0$, $90^0$, $120^0$, $165^0$. Image moments of order up to 15 are used. (solid line: original projections; dash line: projections estimated from Legendre moments; cross: projections estimated from geometric moments)

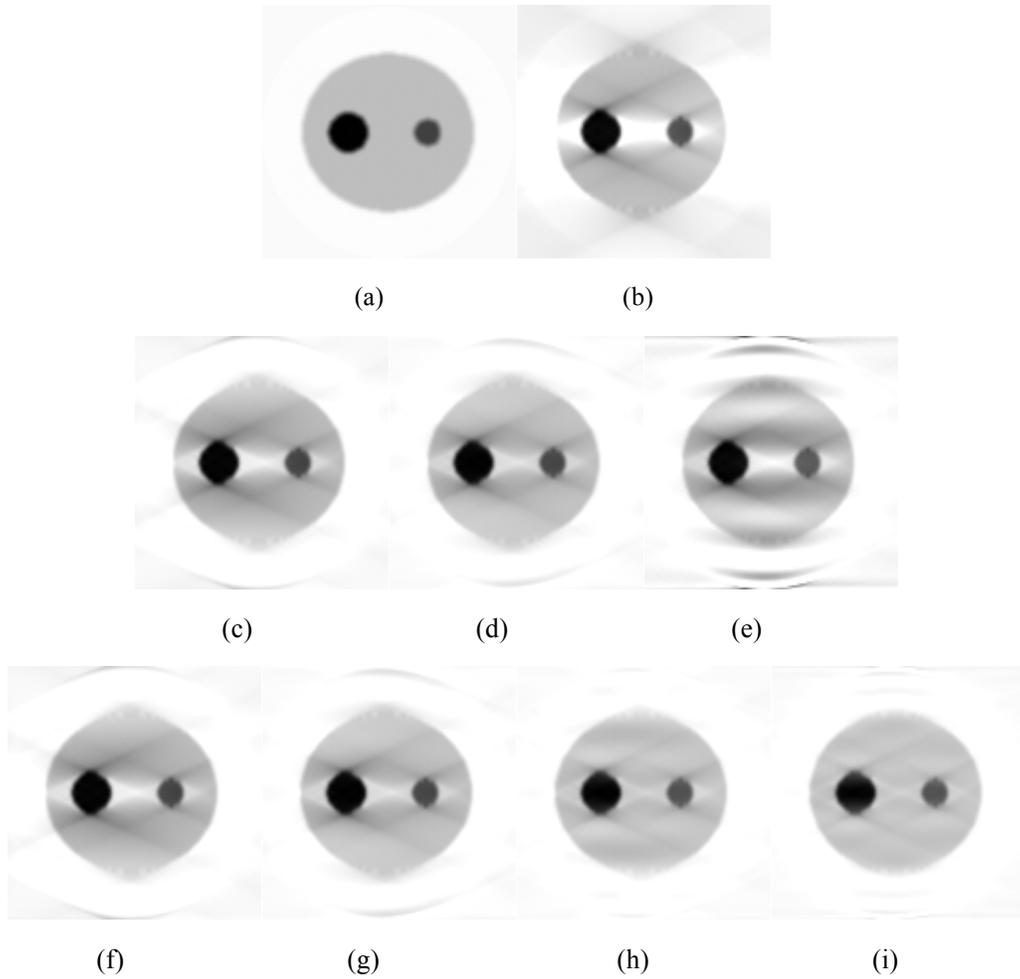

(a)  (b)

(c)  (d)  (e)

(f)  (g)  (h)  (i)

Figure 4. Reconstruction results. (a) Original grey-level image; (b) Reconstruction from incomplete projections ($25^0$-$155^0$) using FBP method; (c), (d) and (e) Reconstructions from incomplete projections using geometric moments with maximum order $M$ = 5, 10, and 15, respectively; (f), (g), (h) and (i) Reconstructions from incomplete projections using orthonormal Legendre moments with maximum order $M$ = 5, 10, 15, and 20, respectively.

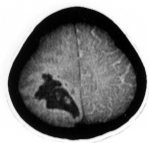

(a)

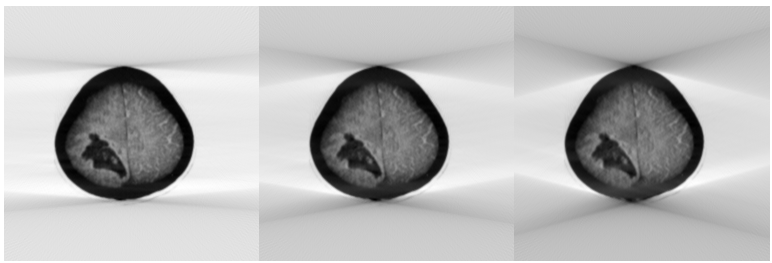

(b)           (c)           (d)

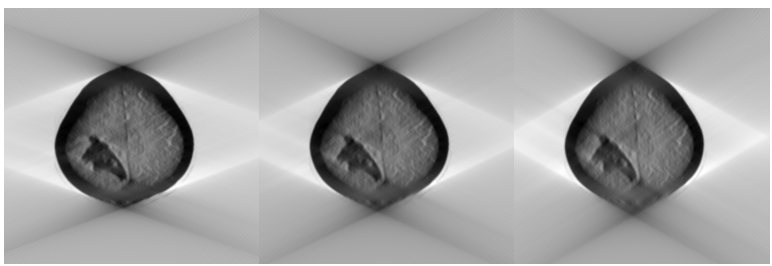

(e)           (f)           (g)

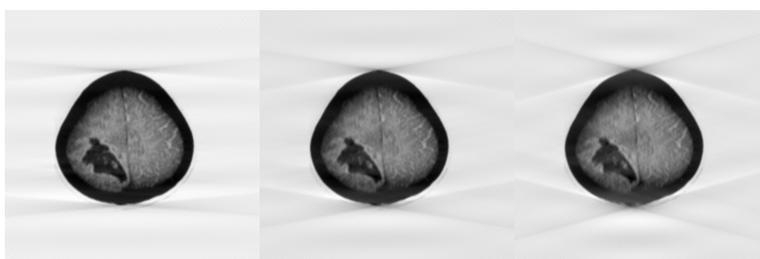

(h)           (i)           (j)

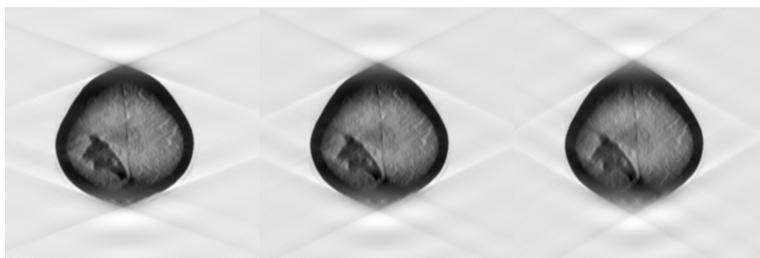

(k)           (l)           (m)

Figure 5. Reconstruction results. (a) Original grey-level image; (b)-(g) Reconstructions from incomplete projections using FBP method for $\alpha$ = 5, 10, 15, 20, 25, and 30, respectively; (h)-(m) Reconstructions from incomplete projections using orthonormal Legendre moments with maximum order $M$ = 25 for $\alpha$ = 5, 10, 15, 20, 25, and 30, respectively.

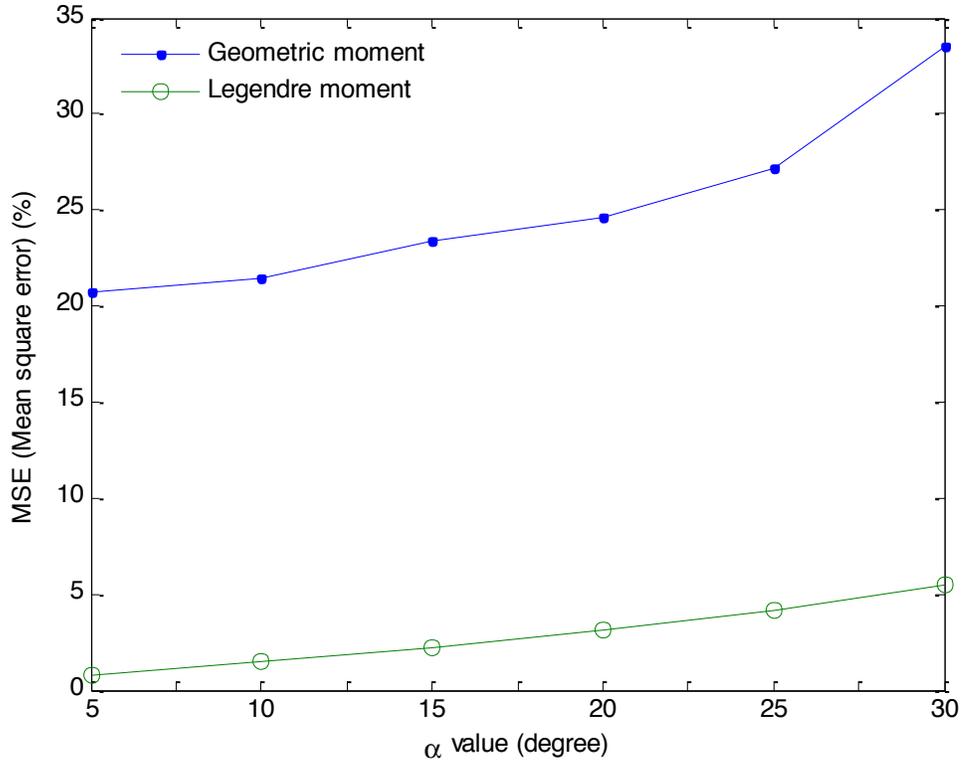

Figure 6. Reconstruction MSE (%) using geometric moment method and orthonormal moment method with maximum order $M$ = 15 for different values of $\alpha$.